%% file: main.tex
\documentclass[sigconf,screen,authorversion]{acmart}

\usepackage{multirow}
\usepackage{anyfontsize} 

\definecolor{gray}{RGB}{111, 111, 111}
\newcommand{\gt}[1]{\textcolor{gray}{#1}}  

\copyrightyear{2025}
\acmYear{2025}
\setcopyright{acmlicensed}

\acmConference[SAC '25]{The 40th ACM/SIGAPP Symposium on Applied Computing}{March 31-April 4, 2025}{Catania, Italy}
\acmBooktitle{The 40th ACM/SIGAPP Symposium on Applied Computing (SAC '25), March 31-April 4, 2025, Catania, Italy}

\acmDOI{10.1145/3672608.3707805}
\acmISBN{979-8-4007-0629-5/25/03}

\begin{document}

\title{Knowledge Distillation in RNN-Attention Models for Early Prediction of Student Performance}
  
\author{Sukrit Leelaluk}
\orcid{0009-0003-3493-6878}
\affiliation{%
  \institution{Kyushu University}
  \department{Graduate School of Information Science and Electrical Engineering}
  \city{Fukuoka} 
  \country{Japan}
}
\email{leelaluksukrit@gmail.com}

\author{Cheng Tang}
\orcid{0000-0002-8148-1509}
\affiliation{%
  \institution{Kyushu University}
  \department{Faculty of Information Science and Electrical Engineering}
  \city{Fukuoka} 
  \country{Japan}
}
\email{tang@ait.kyushu-u.ac.jp}

\author{Valdemar Švábenský}
\orcid{0000-0001-8546-280X}
\affiliation{%
  \institution{Kyushu University}
  \department{Faculty of Information Science and Electrical Engineering}
  \city{Fukuoka} 
  \country{Japan}
}
\email{valdemar.research@gmail.com}

\author{Atsushi Shimada}
\orcid{0000-0002-3635-9336}
\affiliation{%
  \institution{Kyushu University}
  \department{Faculty of Information Science and Electrical Engineering}
  \city{Fukuoka} 
  \country{Japan}
}
\email{atsushi@ait.kyushu-u.ac.jp}


\begin{abstract}
Educational data mining (EDM) is a part of applied computing that focuses on automatically analyzing data from learning contexts. Early prediction for identifying at-risk students is a crucial and widely researched topic in EDM research. It enables instructors to support at-risk students to stay on track, preventing student dropout or failure. Previous studies have predicted students' learning performance to identify at-risk students by using machine learning on data collected from e-learning platforms. However, most studies aimed to identify at-risk students utilizing the entire course data after the course finished. This does not correspond to the real-world scenario that at-risk students may drop out before the course ends. To address this problem, we introduce an RNN-Attention-KD (knowledge distillation) framework to predict at-risk students early throughout a course. It leverages the strengths of Recurrent Neural Networks (RNNs) in handling time-sequence data to predict students' performance at each time step and employs an attention mechanism to focus on relevant time steps for improved predictive accuracy. At the same time, KD is applied to compress the time steps to facilitate early prediction. In an empirical evaluation, RNN-Attention-KD outperforms traditional neural network models in terms of recall and F1-measure. For example, it obtained recall and F1-measure of 0.49 and 0.51 for Weeks 1--3 and 0.51 and 0.61 for Weeks 1--6 across all datasets from four years of a university course. Then, an ablation study investigated the contributions of different knowledge transfer methods (distillation objectives). We found that hint loss from the hidden layer of RNN and context vector loss from the attention module on RNN could enhance the model's prediction performance for identifying at-risk students. These results are relevant for EDM researchers employing deep learning models.
\end{abstract}

\ccsdesc[500]{Computing methodologies~Machine learning}
\ccsdesc[500]{Applied computing~Education}

\keywords{Student Performance Prediction, Educational Data Mining, Learning Analytics, Knowledge Distillation, Neural Network}

\maketitle

\input{body}

\bibliographystyle{ACM-Reference-Format}
\balance
\bibliography{references} 

\end{document}

%% file: body.tex
\section{Introduction}
\label{sec1}

With the development of research fields like Artificial intelligence in education and Educational data mining, many educational institutions are using these methods to enhance e-learning environments~\cite{fu2013, giannakas2021}, for instance, Learning Management Systems (LMS) and Massive Open Online Courses (MOOCs). Students can use these enhanced e-learning platforms to access lecture materials, complete assignments with instant feedback, and communicate with instructors and classmates. Additionally, this allows researchers to collect and use a variety of data to analyze students' learning behaviors, such as demographic data \cite{aissaoui2020}, clicker data from digital textbook systems \cite{chenyang2021, sukrit2022}, quiz scores \cite{okubo2018}, or submission of reports \cite{moises2021}.

Identifying at-risk students has become a crucial topic in educational data mining \cite{hellas2018, hussain2018, koutcheme2022}. At-risk students are more likely to drop out or fail courses, so instructors must provide support to help them stay on track. With the advances in machine learning (ML), many researchers \cite{chenyang2021, he2020, hu2014, kim2018, koutcheme2022, sukrit2022, murata2023, waheed2023a} have investigated predicting students' learning performance to identify at-risk students. To achieve this goal, they employed data from e-learning platforms and used traditional ML or deep learning algorithms \cite{abdullah2023, chenyang2021, feiyue2022, moises2021}. However, most studies applied ML or deep learning models to identify at-risk students by utilizing the whole course data after the course finished \cite{feiyue2022, abdullah2023}. Nevertheless, to increase the applicability of results, identifying at-risk students should be conducted before the course ends to provide timely support, allowing these students to continue in courses with an increased chance of success \cite{zhang2021}.

The present study proposes a novel approach to identifying at-risk students using a deep learning framework called RNN-Attention-KD. We used recurrent neural networks (RNNs), a type of neural network for handling time-series data. We then leveraged the attention mechanism \cite{bahdanau2014, luong2015} to exploit the time sequence of students' learning activities data to improve predictive performance. In addition, we employed knowledge distillation (KD) \cite{hinton2015}. KD is a machine learning technique for compressing model size by transferring knowledge from a large model (known as the \textit{teacher model}) to a smaller model (known as the \textit{student model}). 

We focus on identifying at-risk students early. While students' performance prediction using whole course data generally yields higher accuracy due to the larger dataset, an early prediction that utilizes a limited dataset is more beneficial in actual class settings. This allows for timely interventions to support at-risk students. We thus use KD for time compression in RNNs by transferring knowledge from the teacher model, which was trained on the entire course data with full information, to the student model, which only has information up to the predicted early time step to identify at-risk students. This allows us to utilize only information up to the early weeks in the student model, which performs better than the model trained only on the limited data without knowledge from the teacher model. The contributions of the paper are as follows:
\begin{enumerate}
    \item We introduce KD, which is generally used for model compression applications \cite{hinton2015}, for the novel application of compressing time series to identify at-risk students early using the methods of educational data mining.
    \item We propose the RNN-Attention-KD framework to enhance early performance prediction, which leverages the strengths of RNNs in handling time-sequence educational data to predict students' performance at each time step. The attention mechanism enables the model to focus on relevant time steps for improved predictive accuracy, while KD compresses the time steps to facilitate early prediction.
    \item We analyze and evaluate the performance of RNN-Attention-KD to identify at-risk students early based on the students' learning logs collected from a higher education course.
\end{enumerate}

The rest of this paper is organized as follows: Section \ref{sec2} discusses related studies in the areas of early prediction and KD. Section \ref{sec3} details the RNN-Attention-KD framework. Section \ref{sec4} describes the methods for dataset collection, feature engineering, and evaluation. The empirical results of this study on RNN-Attention-KD are reported in Section \ref{sec5}. Finally, Section \ref{sec6} presents the conclusions and future research challenges stemming from the present study.


\section{Related Work}
\label{sec2}

\subsection{Early Prediction of Student Performance}\label{subsec21}

Predicting student performance is a key research topic in educational data mining and learning analytics, particularly in addressing the issue of at-risk student prediction. With the increasing use and variety of learning management systems (LMS), researchers can collect data on students' learning behaviors and analyze it to develop innovative ways of supporting at-risk students. This helps to prevent them from dropping out or failing their courses \cite{jacob2017, moises2021}.

Many studies used the traditional approach to classifying student performance as at-risk or no-risk, relying on conventional machine learning algorithms such as Logistic Regression (LR) \cite{sagala2022, Svabensky2024detecting}, Support Vector Machine (SVM) \cite{alamri2021, mohd2023, Svabensky2024detecting}, Tree-based algorithms \cite{baradwaj2011, alamri2021, casalino2020, Svabensky2024detecting}, and K-Means Clustering \cite{yann2020, mohd2023}. These algorithms were utilized to identify at-risk students based on stream data within a time sequence. For instance, \citet{casalino2020} processed the Open University Learning Analytics Dataset (OULAD) as a stream of evolving data. OULAD was divided into temporal chunks based on semesters to classify students' learning outcomes using Adaptive Random Forest (ARF). The results showed that ARF efficiently and correctly classified students' outcomes by processing educational data as a stream rather than the non-streamed version, which cannot be reliably used in an actual classroom setting. After deep learning became more prominent in EDM, previous studies employed deep neural networks, such as multi-layer perceptron (MLP) \cite{windarti2020, moises2021}, recurrent neural networks (RNNs) \cite{okubo2017a, vives2024}, and graph neural networks \cite{albreiki2023}, for students' performance prediction tasks.

In addition, \textit{early} prediction of students' learning behaviors is essential for identifying at-risk students as soon as possible. This task aims to recognize and provide early warning to at-risk students who have low performance and tend to drop out, enabling timely intervention by instructors while a course is in progress \cite{hu2014}. Therefore, RNNs algorithms, including the Gated Recurrent Unit (GRU) \cite{cho2014} and the Long Short Term Memory (LSTM) \cite{hochreiter1997}, are the most utilized to predict students' performance using students' learning logs on daily or weekly basis for early prediction \cite{waheed2023a, fatima2023}. For instance, \citet{kim2018} introduced GridNet, which used the bidirectional long short-term memory (Bi-LSTM) as a backbone for early prediction using the data from an online learning platform. They found that GridNet improved student performance prediction accuracy over the weeks. \citet{chenCui2020} utilized LSTM to analyze students' online temporal behaviors on LMS data for early prediction. They compared LSTM with eight conventional machine learning algorithms, measuring the area under the ROC (receiver operating characteristic) curve (AUC) scores. They found that the AUC score of LSTM was 80.1\%, which was higher than those of all the other machine learning classifiers in the first 28 days of data.

\subsection{Knowledge Distillation}\label{subsec22}

Knowledge Distillation (KD) is an ML technique that transfers the knowledge of a \textit{teacher} model to a \textit{student} model while maintaining prediction performance. This is achieved by enabling the student model to mimic the teacher model's behavior by learning its feature representations or reproducing its prediction outputs. 

A loss function generally computes numerical values based on a model's output predictions. In the context of KD, we use the loss function to measure how well the student model's predictions match the target outputs and as a guide for adjusting the student model's internal parameters \cite{gou2021}. However, the loss function itself does not directly update these parameters. We thus rely on a trained teacher model, with fixed parameters, to guide the student model. This encourages the student model to learn representations and behaviors similar to the teacher model.

The KD method was first proposed by \citet{hinton2015}'s study, where knowledge is distilled from the teacher model by minimizing the Kullback-Leibler divergence between the teacher's soft logits and the student's predictions. Subsequently, many KD methods that leverage intermediate features from the teacher model to propagate knowledge to the student model were proposed to enhance the student model's prediction performance. FitNets \cite{romero2015}, a feature-based distillation method, introduced the concept of transferring knowledge by minimizing the Euclidean distance between intermediate layers of the teacher and student models. Attention distribution transfer \cite{zagoruyko2017, sungho2022} was introduced to propagate knowledge from the attention maps of a convolutional network, which represent specific features of the teacher model, to the student model.

KD is typically used to compress models for specific tasks such as language modelling \cite{jiao2020, wang2021}, low-resolution image recognition~\cite{sungho2022}, and object detection \cite{chen2017a, wei2022}. In this study, we apply the KD method to our model to achieve high efficiency in early prediction tasks for detecting at-risk students. Data about students' learning activity is time-sequential, as their behaviors change weekly based on class activities and their cognition development, which typically improves after each class. We thus utilize RNNs to handle the time-series data and KD for time-series compression, aiming to enhance early prediction capabilities \cite{murata2023}. 

To the best of our knowledge, \citet{murata2023} was the first to propose RNN-FitNets. They utilized RNNs to learn the time-series information inherent in student learning behavior data. FitNets~\cite{romero2015} reproduced the hidden state of the RNN in a student model, which was trained with a teacher model using whole-course data, enabling compressed time-series representation for early prediction. 

However, a significant shortcoming of RNNs is that they may be prone to the problem of vanishing gradients when processing long input sequences, which means the model struggles to properly learn patterns from earlier time steps during training. This happens because the model tends to forget earlier inputs as a model moves through the subsequent parts of the sequence~\cite{bengio1994}. This issue is particularly concerning in early prediction tasks, where information from the initial weeks may be crucial. To address this, we propose an RNN with an attention mechanism that retains essential information from the entire input sequence by focusing on critical parts of the time series \cite{bahdanau2014, luong2015}. This enhanced focus can also improve the guidance provided to the student model during KD from the teacher model \cite{wang2022}, which our study aims to evaluate.


\section{Research Methods}
\label{sec3}

\subsection{Problem Statement}\label{subsec31}

The problem statement for early prediction of at-risk students using KD can be defined as follows: Let $D = \{d_1, d_2, \ldots, d_K\}$ represent the dataset containing the learning activity data of all students in a course, where $d_k$ denotes the data for student $k$. For each student $k$, $d_k$ comprises the learning activity features collected weekly over the course duration of $M$ weeks, expressed as $d_k = \{(x_1, y_1), (x_2, y_2), \ldots, (x_M, y_M)\}$. Here, $x_i$ represents the students' learning activity features of a week $i$ and $y_i$ is the corresponding at-risk or non-at-risk label.

The teacher model $f_T : X_M \to Y$ is trained on the complete course dataset $D$, where $X_M$ is the input space of features from all $M$ weeks. The objective of our study is to design a KD framework to train a student model $f_S : X_N \to Y$ using $D^{\text{early}}$ and knowledge transferred from $f_T$. Here, $d_k^{\text{early}} = \{(x'_1, y_1), (x'_2, y_2), \ldots, (x'_N, y_N)\}$ is a subset of data for student $k$, where $x'_i$ represents the learning features of student $k$ for only the first $N$ weeks ($N < M$).

Our study aims to find the optimal parameters for $f_S$ that enable accurate predictions and identification of at-risk students using only early-week data, while leveraging the knowledge distilled from the more comprehensive teacher model $f_T$. The objective function can be expressed as Equation \ref{eq_objective}:

\begin{equation}
\operatorname{argmin}_{f_S, f_T} L(f_S(x'_i), y_{\text{true}}, f_T(x_i)),
\label{eq_objective}
\end{equation}
where $L$ is a total loss function that incorporates both the true labels $y_{\text{true}}$ and the soft targets provided by the teacher model $f_T$, $f_S(x'_i)$ is the prediction of the student model on early-week data, and $f_T(x_i)$ is the prediction of the teacher model on full-course data. The total loss function $ L $ is further described in Section \ref{subsec32}.

\subsection{Overview of the Proposed Framework}\label{subsec32}

This section introduces the architecture of the RNN with an attention mechanism model (RNN-Attention), which serves as the teacher model in the KD framework. Subsequently, we describe the KD method proposed in this study, which aims to compress the time series length and detect at-risk students at an early stage while maintaining high prediction performance.

\subsubsection{RNN with an Attention Mechanism Model (RNN-Attention)}
\label{subsubsec321}

\begin{figure}[t!]
    \centering
    \includegraphics[width=0.65\linewidth]{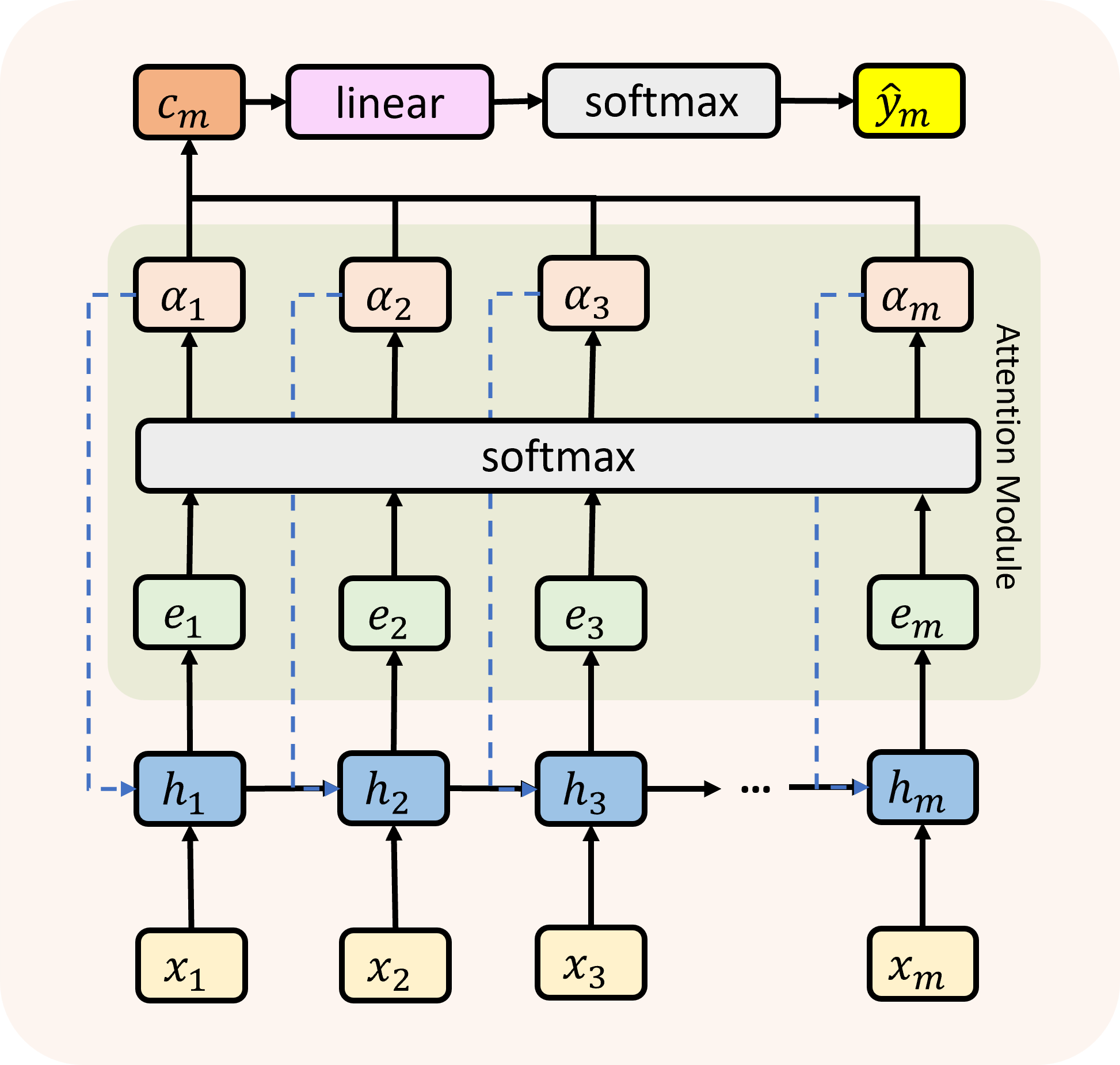}
    \caption{\textbf{RNN with an attention mechanism structure (RNN-Attention)}.}
    \label{fig:rnn-attention}
\end{figure}

In this study, we employ the RNN model shown in Fig. \ref{fig:rnn-attention}, a type of neural network, to process students' learning activity data, which were collected sequentially each week throughout a course. RNN has a loop structure in its hidden layer that enables it to incorporate information from past and current inputs. For prediction, RNN receives an input sequence $ X = (x_1, x_2, ..., x_m) $ with \( x_i \in \mathbb{R}^k \) from the first to the $ m^{\text{th}} $ time steps, where $ k $ represents the students' learning activity features. The hidden state $ h_i $ is calculated as follows:

\begin{equation}
\mathbf{h}_i = f(\mathbf{h}_{i-1}, \mathbf{x}_i)
\label{eq_rnn_basic}
\end{equation}
Here, \( h_i \in \mathbb{R}^r \) is the hidden state at the $ i^{\text{th}} $ time step, $ r $ denotes the size of the hidden state, and $ f $ is a non-linear activation function within the RNN. However, RNNs are prone to vanishing gradient problems \cite{bengio1994}. Therefore, $ f $ could alternatively be an LSTM \cite{hochreiter1997} or GRU \cite{cho2014} activation function.

In natural language processing (NLP), RNNs, including LSTM and GRU, are commonly employed in machine translation as part of an encoder-decoder framework \cite{sutskever2014}. In this approach, the source sentence is encoded into a fixed-length vector, which is then passed to the decoder to generate the translation. However, a significant limitation arises when the source sentence is lengthy: the model tends to forget earlier inputs as it processes subsequent parts of the sequence. This issue is particularly concerning in early prediction tasks, where information from the initial stages may be crucial for identifying at-risk individuals. To address this problem, the attention mechanism is integrated with RNNs to ensure the model retains and utilizes important information from the entire sequence.

After the hidden layers from RNNs, the output vector from the hidden layers serves as input to the attention module, where the attention mechanism calculates the attention weight $ a_i $ for each time step $ i $. Then, the context vector $ C_m $ is obtained from the summation of the product between each attention weight $ a_i $ and the output from hidden layer $ h_i $ from RNNs.

In the traditional encoder-decoder model, the attention score is typically calculated based on the alignment score between the hidden state of each time step on the encoder and the current time step on the decoder. However, in our proposed model, which utilizes a single RNN structure, the alignment score $ e_i $ is computed between the hidden state of each time step $ h_i $ and the final time step $ h_m $, as described in Eq. (\ref{eq_alignment_score}). The final hidden $ h_m $ of the RNN contains the most comprehensive information that is accumulating information from each step throughout the sequence \cite{xie2019}.

\begin{equation}
e_i = (\mathbf{h}_m)^\top \mathbf{W} \mathbf{h}_i
\label{eq_alignment_score}
\end{equation}
Here, \( e_i \in \mathbb{R}^1 \) is the alignment score of the time step $ i^{\text{th}} $, \( h_m \in \mathbb{R}^r \) is the hidden state on the final time step, and \( W \in \mathbb{R}^{r \times r} \) is a trainable parameter. Then, a softmax function is applied to the alignment score $ e_i $ to ensure the attention weight $ \alpha_i $ sums to 1, as in Eq. (\ref{eq_attention_score}).

\begin{equation}
\alpha_i = \frac{\exp(e_i)}{\sum_{j=1}^{m} \exp(e_j)}
\label{eq_attention_score}
\end{equation}

Subsequently, we employed the attention scores $ \alpha_i $ (representing probability information) associated with the original hidden states $ h_i $ at each time step to compute the context vector \( C_m \in \mathbb{R}^r \) as a weighted sum of the hidden states from all time steps.

\begin{equation}
C_m = \sum_{i=1}^{m} \alpha_i \mathbf{h}_i
\label{eq_context_vectors}
\end{equation}

Once the context vector $ C_m $ is obtained from the attention module, we employ it as input to a multi-layer perceptron (MLP) for predicting the final output $ \hat{y}_m $. MLP comprises hidden units with the same dimension as the RNN's hidden size and two output units for classifying the probability between at-risk and no-risk.

\subsubsection{Knowledge Distillation Framework}
\label{subsubsec322}

\begin{figure*}[t!]
    \centering
    \includegraphics[width=0.65\linewidth]{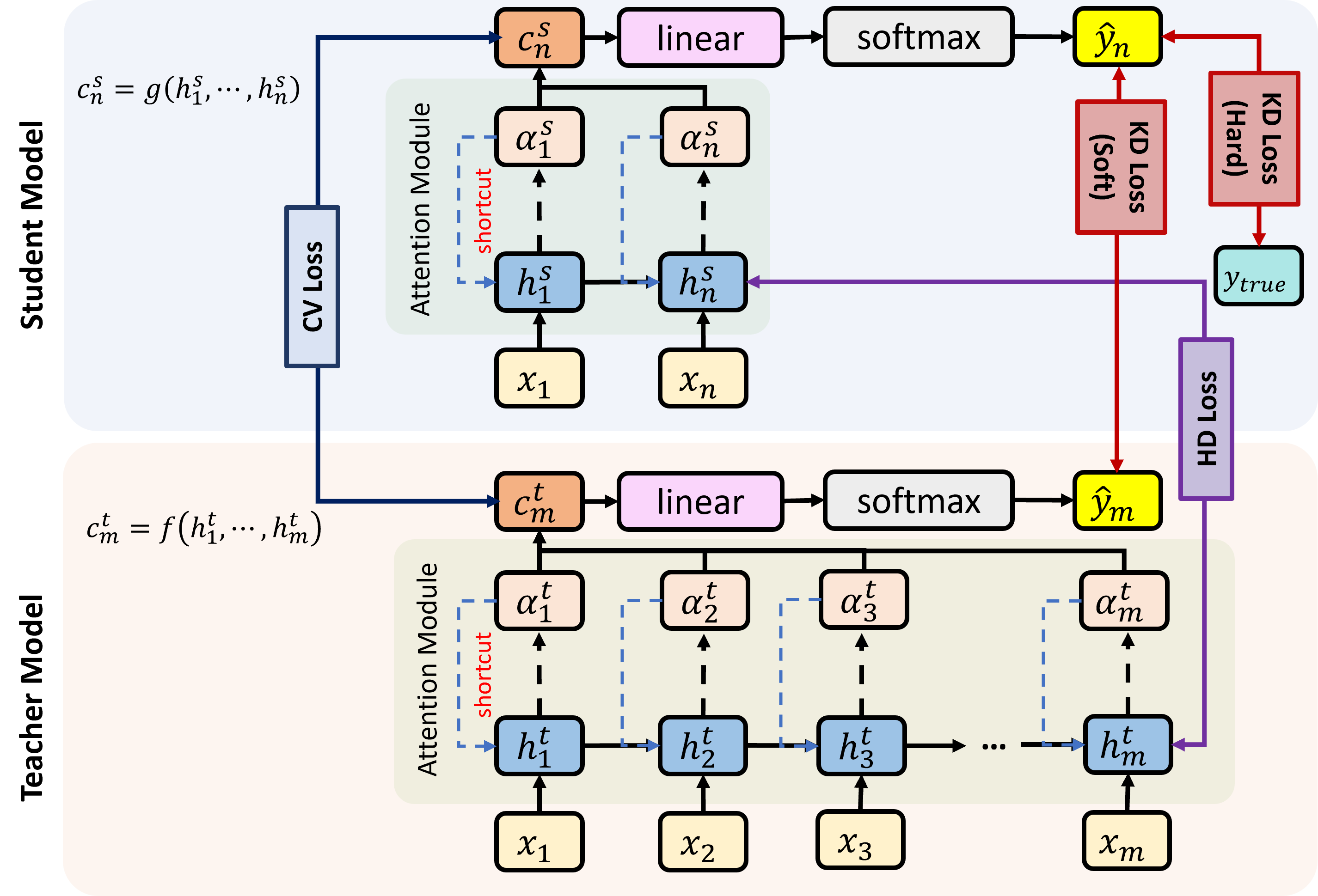}
    \caption{\textbf{Knowledge Distillation Framework of the RNN with an attention mechanism structure (RNN-Attention-KD)}.}
    \label{fig:rnn-attention-kd}
\end{figure*}

In this section, we propose a KD method for RNN with an attention mechanism (RNN-Attention-KD) on early prediction tasks. We employ the KD method for time series compression to reproduce a hidden state of the RNN model, which was trained with a teacher model using whole-course data. Additionally, the attention mechanism is leveraged to enhance performance by guiding the student model to focus on the critical parts of the input time sequence. Figure \ref{fig:rnn-attention-kd} shows an overview of RNN-Attention-KD.

The teacher model $ t $ and student model $ s $ have structure as described in Section \ref{subsubsec321}, including the input feature, number of layers, and number of hidden units. The teacher model is pre-trained using students' learning activity data from the entire course, spanning all time steps (1, 2, \dots, $ m $). In contrast, the student model is trained up to the $n^{\text{th}}$ time step, where $1 \leq n \leq m$. Therefore, the teacher and student models differ in their time series lengths.

The student model is trained in three steps during each epoch. First, the student model updates its parameters through the hidden layers, excluding the attention modules and output nodes. The parameters of the hidden layers are updated by minimizing the hint loss function $ \mathcal{L}_{HD} $. This allows the student model to learn to mimic the teacher model’s hidden states, capturing the rich time series information from the entire course. The objective is defined as:

\begin{equation}
\mathcal{L}_{HD} = \text{MSE}(h_m^t, h_n^s),
\label{loss_hidden}
\end{equation}
where \( h_m^t \in \mathbb{R}^r \) and \( h_n^s \in \mathbb{R}^r \) refer to the hidden state of the teacher and student model, respectively. Since the structure of teacher and student models are the same, the number of hidden nodes $ r $ are also the same. MSE is the mean squared error loss function.

Once the hint loss function is updated, the attention module parameters within the student model will be updated next. Adjusting the attention weights on the context vector $ C $ enables the student model to capture valuable information about the relationship between early and future learning sessions, which is necessary for identifying at-risk students. We thus propose a context vector loss function (attention-based distillation) $ \mathcal{L}_{CV} $ to encourage knowledge transfer from the teacher to the student network. The objective of this function is as follows:

\begin{equation}
\mathcal{L}_{CV} = \text{MSE}(c_m^t, c_n^s),
\label{loss_context_vector}
\end{equation}
where \( c_m^t \in \mathbb{R}^r \) and \( c_n^s \in \mathbb{R}^r \) refer to the teacher and student model context vector, respectively. 

Finally, the entire student model, including the output layers, is updated by minimizing the distillation loss $ \mathcal{L}_{KD} $. We apply two penalties: the soft cross-entropy loss between the predicted logits of the student and teacher models and the hard cross-entropy loss between the predicted logits of the student model and the ground truth. Soft cross-entropy measures how well the student model's predictions match the teacher model's predictions. However, the teacher model has the possibility of not providing the correct ground truth in some cases. Therefore, hard cross-entropy gauges how the student model's predictions match the ground truth. This ensures that the student model learns to make accurate predictions independently, beyond just copying the teacher model's predictions. The objective of this function is as follows:

\begin{equation}
\mathcal{L}_{KD} = \mathcal{H}(\mathbf{y}_{\text{true}}, \hat{y}_n) + \lambda \mathcal{H}(\hat{y}_m, \hat{y}_n),
\label{loss_distillation}
\end{equation}
where $ \mathcal{H} $ is the cross-entropy loss, $ \lambda $ is a hyperparameter that balances soft and hard cross-entropy losses, $ \mathbf{y}_\text{true} $ is a true label, and $ \hat{y}_m $ and $ \hat{y}_n $ are predicted logits from the teacher and student model, respectively.


\section{Experimental Setting}
\label{sec4}

\subsection{Dataset and Feature Engineering}\label{subsec41}

The study utilized data collected from a programming theory (PT) course, a required subject for second-year undergraduate students at the Faculty of Information Science and Engineering, Kyushu University, Japan. In the course, students learn basic knowledge and skills of computer programming by using the Scheme language. The course attendees are beginners in this programming language. 

Data of the total of 219 students throughout the spring quarters from 2019 to 2022 were used. Each quarter lasted seven weeks. All four course offerings employed the same syllabus, although the teaching modality differed: PT2019 was held on-site, and the other three were held online due to the COVID-19 pandemic restrictions. 

The grade distribution in the PT courses is listed in Table \ref{table_grade}. Due to the sometimes limited number of students receiving grade points of C, D, and F, our study employed a binary classification approach rather than a five-label classification. Students were categorized into two groups: at-risk and non-at-risk. Those who obtained grade points of C, D, or F were classified as at-risk, while students achieving higher grades were designated as non-at-risk.

\begin{table}[h]
\caption{Distribution of grades per number of students (219 students total) in the Programming Theory (PT) courses.}
\label{table_grade}
\begin{tabular*}{\columnwidth}{@{\extracolsep\fill}l|rr|rr|rr|rr}
\toprule%
Grade & \multicolumn{2}{c|}{PT2019} & \multicolumn{2}{c|}{PT2020} & \multicolumn{2}{c|}{PT2021} & \multicolumn{2}{c}{PT2022} \\
\midrule
A & 24 & \gt{(48.0\%)} & 22 & \gt{(35.5\%)} & 20 & \gt{(37.0\%)} & 17 & \gt{(32.1\%)} \\
B &  6 & \gt{(12.0\%)} & 24 & \gt{(38.7\%)} & 15 & \gt{(27.8\%)} &  7 & \gt{(13.2\%)} \\
C &  4 &  \gt{(8.0\%)} &  6 &  \gt{(9.7\%)} & 10 & \gt{(18.5\%)} &  5 &  \gt{(9.4\%)} \\
D &  6 & \gt{(12.0\%)} &  3 &  \gt{(4.8\%)} &  3 &  \gt{(5.6\%)} & 22 & \gt{(41.5\%)} \\
F & 10 & \gt{(20.0\%)} &  7 & \gt{(11.3\%)} &  6 & \gt{(11.1\%)} &  2 &  \gt{(3.8\%)} \\ 
\hline
Total & 50 &  & 62 &  & 54 &  & 53 &  \\ 
\bottomrule
\end{tabular*}
\end{table}

Students used Moodle, a learning management system, and BookRoll \cite{ogata2015}, an e-book system, to access lecture materials and assignments. Both of these systems automatically logged student interactions, which were then converted to features as shown in Table \ref{table_description}. Attendance and report submission data were extracted from Moodle, categorizing student activities for each lecture. Additional data, including reading time, notes, total actions, and others were obtained from BookRoll. Combining these data sources enabled us to more accurately capture student learning progressions.

\begin{table*}[ht]
\caption{Description of each feature for prediction models based on student activities.}
\label{table_description}
\begin{tabular}{lll}
\hline
Platforms & Activities & Explanation \\ \hline
\multirow{3}{*}{LMS (Moodle)}      & Attendance          & Tracks attendance (present, late, absent) for each lecture             \\
                                   & Report              & Tracks report submission status (on time, late, none) for each lecture \\
                                   & Course Accesses     & Total number of course accesses on Moodle                              \\ \hline
\multirow{9}{*}{E-Book (BookRoll)} & Reading Time        & Reading time of lecture materials (in seconds)                            \\
                                   & Markers             & Number of markers (highlights) a student added                         \\
                                   & Memos               & Number of memos (notes) a student added                                \\
                                   & Total Actions       & Total number of reading operation events                               \\
                                   & Open                & Number of times a student opened the e-book                            \\
                                   & Next                & Number of times a student went to the next page                        \\
                                   & Prev                & Number of times a student backed to the previous page                  \\
                                   & Page Jump           & Number of times a student jumped to a particular page                  \\
                                   & Close               & Number of times a student closed the e-book                            \\ \hline
\end{tabular}
\end{table*}

The feature values were determined based on Score Ranking Points (SRPs). SRPs, inspired by Active Learner Points (ALPs)~\cite{okubo2017a, okubo2017b, okubo2018}, are an evaluation system that assesses students' learning activities through weekly scores based on percentile rankings within a given class. Specifically, scores from 10 to 1 were incorporated, as shown in Table \ref{table_srp}, to enhance the feature vector's representational capacity and facilitate the model's comprehension of underlying patterns. This refined scale offers greater granularity compared to the 5 to 1 scale used in ALPs. Consequently, twelve SRPs were incorporated as input features for the prediction model. In cases where multiple report assignments were due within a single week, the report scores were averaged. Before the SRPs were input into the prediction model, all SRPs were normalized to a 0–1 range to ensure consistency and facilitate training.

\begin{table}[ht]
\caption{Criteria for Score Ranking Points (SRPs).}
\label{table_srp}
\begin{tabular}{ll}
\hline
Feature & Active Score \\ \hline
Attendance & \begin{tabular}[c]{@{}l@{}}Attendance = 10, Late = 5, \\ Absence = 0\end{tabular} \\ \hline
Report & \begin{tabular}[c]{@{}l@{}}On Time = 10, Late = 5, \\ None = 0
\end{tabular} \\ \hline
\begin{tabular}[c]{@{}l@{}}Course Accesses, Reading Time, \\ Markers,  Memos, Total Actions, \\ Open, Next, Prev, Page Jump, Close\end{tabular} & \begin{tabular}[c]{@{}l@{}}Top 1--10\% = 10, \\ Top 11--20\% = 9, ..., \\ Bottom 91--100\% = 1\end{tabular} \\ \hline
Any feature with zero activity & 0 score \\ \hline
\end{tabular}
\end{table}

In order to simulate the real-world scenario of forecasting future course outcomes based on previously available data, four PT courses were divided into six datasets. Each dataset represents a task where training is performed on one or more prior-year courses, while the subsequent year’s course serves as the test set. This approach enables a more robust evaluation of model performance, ensuring that the results are not tied to a single dataset but remain consistent across multiple datasets. The datasets are detailed in Table \ref{table_dataset}. T and P denote train and test (predict) datasets, respectively. The T19P20, T20P21, and T21P22 datasets each consisted of one training course and one for testing from the following year. The T1920P21 and T2021P22 dataset comprised two consecutive years of courses used for training, while the T192021P22 dataset comprised three. In both cases, one course from the subsequent year was reserved for testing.

\begin{table}[h]
\caption{The details of the datasets.}
\label{table_dataset}
\begin{tabular}{@{}lll@{}}
\toprule
Dataset & Train \& Validation course & Test Course \\
\midrule
T19P20 & 2019 & 2020 \\
T20P21 & 2020 & 2021 \\
T21P22 & 2021 & 2022 \\
T1920P21 & 2019, 2020 & 2021 \\
T2021P22 & 2020, 2021 & 2022 \\
T192021P22 & 2019, 2020, 2021 & 2022 \\
\bottomrule
\end{tabular}
\end{table}

\subsection{Evaluation Methods}\label{subsec42}

\subsubsection{Metrics}
\label{subsubsec421}

This study employed three metrics to evaluate the performance of the binary classifiers: precision, recall, and F1-measure. Given the imbalanced nature of the experimental data, weighted average F1-scores were calculated across multiple runs. A confusion matrix was utilized to quantify true positive (TP), false positive (FP), false negative (FN), and true negative (TN) instances in the test samples to compute precision, recall, and F1-measure. We assigned at-risk students a label of 1 (positive class) and non-at-risk students a label of 0 (negative class), following the standard binary classification convention~\cite{Svabensky2024detecting}. Precision is calculated as the ratio between the number of correctly predicted at-risk students and the total number of students predicted as at-risk, whereas recall is defined as the ratio between the number of correctly predicted at-risk students and the total number of actual at-risk students. The F1-score, which is calculated from the harmonic mean of precision and recall, represents both precision and recall in one metric.

\subsubsection{Selection of Models and Parameters}
\label{subsubsec422}

For the hyperparameter search (using grid search) on each dataset, we employed 5-fold cross-validation on the train set to find the optimal hyperparameters on RNN-Attention-KD. We adjusted the backbone models, GRU and LSTM. Since the data size is limited, we use one hidden layer on GRU and LSTM, with the following units: 4, 6, 8, and 10. We then adjusted the learning rate with 0.01 and 0.001, with weight decay (L2 regularization) of 1E-5, batch size of 8, and epochs at 150. 

Based on the average cross-validation performance across all datasets, we selected the GRU as the backbone, with one hidden layer and four units. The GRU is more effective at handling dependencies in shorter time series, whereas the LSTM performs better with longer time series \cite{he2020}. Since our datasets consist of weekly units, each dataset spanning seven weeks, the GRU is more suitable for this case. After the hyperparameter search, the prediction model was trained using the entire training set, and the result was subsequently evaluated on test set.


\section{Experimental Results and Discussion}
\label{sec5}

In this study, we first compare the ability to detect at-risk students using conventional algorithms, followed by an ablation study to investigate the contribution of different distillation objectives in RNN-Attention-KD. When testing the performance of models, we executed the training independently 30 times, and the metrics' means were compared in all executions to avoid stochastic results.

\subsection{Ability to Detect At-risk Students}
\label{subsec51}

\begin{table*}[t]
\centering
\caption{Comparison of the ability for detecting at-risk students. The results report Precision (PR), Recall (RE), and F1-measure (F1). Bold numbers indicate the highest value for each metric in each week.}
\label{table_expt1}
\resizebox{\textwidth}{!}{%
\begin{tabular}{|cc|ccc|ccc|ccc|ccc|ccc|ccc|ccc|}
\hline
\multirow{2}{*}{Dataset}    & \multirow{2}{*}{Week (s)} & \multicolumn{3}{c}{MLP} & \multicolumn{3}{c}{RNN} & \multicolumn{3}{c}{GRU} & \multicolumn{3}{c}{LSTM} & \multicolumn{3}{c}{Bi-GRU} & \multicolumn{3}{c}{Bi-LSTM} & \multicolumn{3}{c|}{RNN-Attention-KD} \\ \cline{3-23} 
                            &                           & PR     & RE     & F1    & PR     & RE     & F1    & PR     & RE     & F1    & PR     & RE     & F1     & PR      & RE      & F1     & PR      & RE      & F1      & PR     & RE     & F1     \\ \hline
\multirow{4}{*}{T19P20}     & 1-3                       & 0.44   & \textbf{0.84}& \textbf{0.58}& \textbf{0.47}& 0.76   & \textbf{0.58}& 0.45   & 0.73   & 0.56  & 0.45   & 0.78   & 0.57   & 0.46    & 0.72    & 0.56   & 0.45    & 0.81    & 0.57    & 0.41   & 0.83   & 0.55   \\
                            & 1-4                       & 0.47   & 0.69   & 0.56  & 0.50   & 0.70   & 0.59  & 0.51   & 0.71   & 0.59  & 0.52   & 0.75   & 0.61   & \textbf{0.54}& 0.72    & 0.62   & \textbf{0.54}& 0.76    & \textbf{0.63}& 0.48   & \textbf{0.81}& 0.60   \\
                            & 1-5                       & 0.54   & 0.68   & 0.60  & 0.62   & 0.73   & \textbf{0.67}& 0.62   & 0.71   & 0.66  & 0.59   & \textbf{0.75}& 0.66   & \textbf{0.65}& 0.69    & 0.67   & 0.63    & 0.73    & \textbf{0.67}& 0.62   & 0.72   & 0.66   \\
                            & 1-6                       & 0.58   & 0.70   & 0.63  & 0.66   & 0.77   & 0.71  & \textbf{0.68}& 0.80   & \textbf{0.73}& 0.65   & 0.82   & 0.72   & 0.65    & 0.79    & 0.71   & 0.64    & \textbf{0.83}& 0.72    & 0.66   & 0.79   & 0.72   \\ \hline
\multirow{4}{*}{T20P21}     & 1-3                       & 0.53   & 0.33   & 0.39  & 0.89   & 0.30   & 0.44  & 0.89   & 0.29   & 0.43  & \textbf{0.94}& 0.31   & 0.46   & 0.89    & 0.28    & 0.42   & 0.89    & 0.35    & 0.49    & 0.56   & \textbf{0.47}& \textbf{0.50}\\
                            & 1-4                       & 0.48   & 0.28   & 0.36  & 0.73   & 0.34   & 0.46  & 0.72   & 0.35   & 0.47  & \textbf{0.83}& 0.39   & \textbf{0.53}& 0.71    & 0.34    & 0.46   & 0.69    & 0.36    & 0.48    & 0.59   & \textbf{0.44}& 0.50   \\
                            & 1-5                       & 0.70   & 0.38   & 0.49  & 0.80   & 0.41   & 0.53  & 0.81   & 0.44   & 0.57  & 0.80   & 0.46   & 0.58   & 0.84    & \textbf{0.47}& \textbf{0.60}& \textbf{0.87}& 0.45    & 0.59    & 0.71   & \textbf{0.47}& 0.55   \\
                            & 1-6                       & 0.74   & 0.39   & 0.51  & 0.79   & 0.49   & 0.60& \textbf{0.87}& 0.48   & 0.61  & 0.81   & 0.44   & 0.57   & 0.81    & 0.45    & 0.58   & 0.78    & 0.48    & 0.59    & 0.83   & \textbf{0.52}& \textbf{0.64}\\ \hline
\multirow{4}{*}{T21P22}     & 1-3                       & 0.58   & 0.31   & 0.40  & 0.82   & 0.30   & 0.44  & 0.83   & 0.29   & 0.42  & 0.81   & 0.32   & 0.46   & \textbf{0.90}& 0.31    & 0.46   & 0.75    & 0.34    & 0.46    & 0.67   & \textbf{0.40}& \textbf{0.50}\\
                            & 1-4                       & 0.57   & 0.29   & 0.39  & \textbf{0.91}& 0.35   & 0.50  & 0.89   & 0.33   & 0.48  & 0.84   & 0.37   & \textbf{0.51}& 0.86    & 0.33    & 0.48   & 0.81    & 0.36    & 0.50    & 0.69   & \textbf{0.41}& \textbf{0.51}\\
                            & 1-5                       & 0.60   & 0.29   & 0.39  & \textbf{0.91}& 0.37   & 0.52  & \textbf{0.91}& 0.35   & 0.51  & 0.85   & 0.39   & 0.53   & 0.90    & 0.37    & 0.52   & 0.83    & 0.41    & 0.55    & 0.82   & \textbf{0.44}& \textbf{0.57}\\
                            & 1-6                       & 0.74   & 0.27   & 0.40  & \textbf{0.94}& 0.38   & 0.53  & 0.93   & 0.39   & 0.55  & 0.87   & 0.38   & 0.53   & \textbf{0.94}& 0.39    & 0.55   & 0.87    & 0.37    & 0.52    & 0.86   & \textbf{0.42}& \textbf{0.56}\\ \hline
\multirow{4}{*}{T1920P21}   & 1-3                       & 0.49   & 0.36   & 0.41  & 0.80   & \textbf{0.37}& 0.49  & 0.85   & 0.36   & \textbf{0.50}& \textbf{0.87}& 0.33   & 0.47   & 0.78    & 0.36    & 0.49   & 0.84    & 0.31    & 0.45    & 0.61   & \textbf{0.37}& 0.45   \\
                            & 1-4                       & 0.55   & 0.34   & 0.41  & 0.75   & 0.42   & 0.54& 0.78   & 0.46   & 0.57  & 0.81   & 0.43   & 0.56   & 0.80    & 0.44    & 0.57   & \textbf{0.84}& 0.41    & 0.55    & 0.72   & \textbf{0.49}& \textbf{0.58}\\
                            & 1-5                       & 0.78   & 0.39   & 0.52  & 0.80   & 0.45   & 0.58  & 0.84   & 0.45   & 0.58  & 0.92   & 0.45   & 0.60   & 0.84    & 0.48    & \textbf{0.61}& \textbf{0.97}& 0.42    & 0.59    & 0.74   & \textbf{0.50}& 0.59   \\
                            & 1-6                       & 0.82   & 0.46   & 0.59  & 0.84   & \textbf{0.48}& \textbf{0.61}& 0.83   & 0.45   & 0.58  & 0.88   & 0.46   & 0.60   & 0.85    & 0.45    & 0.59   & \textbf{0.91}& 0.45    & 0.60    & 0.79   & \textbf{0.48}& 0.59   \\ \hline
\multirow{4}{*}{T2021P22}   & 1-3                       & 0.69   & 0.30   & 0.41  & 0.78   & 0.36   & 0.49  & 0.78   & 0.34   & 0.47  & 0.77   & 0.34   & 0.47   & \textbf{0.83}& 0.36    & 0.50   & 0.77    & 0.38    & 0.50    & 0.68   & \textbf{0.42}& \textbf{0.52}\\
                            & 1-4                       & 0.64   & 0.36   & 0.45  & 0.80   & 0.37   & 0.50  & \textbf{0.81}& 0.39   & 0.52  & 0.79   & 0.40   & 0.53   & 0.78    & 0.36    & 0.49   & 0.74    & 0.39    & 0.51    & 0.72   & \textbf{0.46}& \textbf{0.56}\\
                            & 1-5                       & 0.74   & 0.40   & 0.52  & \textbf{0.87}& 0.42   & 0.56  & 0.86   & 0.42   & 0.56  & 0.84   & 0.41   & 0.55   & \textbf{0.87}& 0.38    & 0.53   & 0.85    & 0.45    & \textbf{0.59}& 0.82   & \textbf{0.47}& \textbf{0.59}\\
                            & 1-6                       & 0.84   & 0.39   & 0.53  & 0.86   & 0.39   & 0.53  & \textbf{0.88}& 0.35   & 0.50  & 0.83   & 0.39   & 0.53   & 0.87    & 0.35    & 0.49   & 0.82    & \textbf{0.41}& \textbf{0.54}& 0.82   & 0.39   & 0.52   \\ \hline
\multirow{4}{*}{T192021P22} & 1-3                       & 0.71   & 0.33   & 0.44  & \textbf{0.90}& 0.41   & 0.56  & 0.88   & 0.40   & 0.55  & 0.89   & 0.41   & 0.56   & 0.87    & 0.41    & 0.56   & 0.86    & 0.42    & \textbf{0.57}& 0.74   & \textbf{0.43}& 0.54   \\
                            & 1-4                       & 0.78   & 0.36   & 0.49  & \textbf{0.85}& 0.41   & \textbf{0.55}& 0.82   & 0.39   & 0.52  & 0.81   & 0.39   & 0.53   & 0.82    & 0.36    & 0.49   & 0.76    & 0.35    & 0.48    & 0.76   & \textbf{0.43}& \textbf{0.55}\\
                            & 1-5                       & 0.82   & 0.39   & 0.52  & 0.87   & 0.44   & 0.58  & 0.87   & 0.45   & 0.59  & 0.86   & 0.45   & 0.59   & \textbf{0.88}& 0.47    & 0.61   & 0.86    & 0.47    & 0.61    & 0.83   & \textbf{0.54}& \textbf{0.65}\\
                            & 1-6                       & \textbf{0.93}& 0.39   & 0.55  & 0.87   & 0.43   & 0.57  & 0.85   & 0.42   & 0.57  & 0.85   & 0.41   & 0.55   & 0.86    & 0.43    & 0.58   & 0.83    & 0.43    & 0.56    & 0.81   & \textbf{0.48}& \textbf{0.60}\\ \hline
\end{tabular}%
}
\end{table*}

\begin{table*}[t]
\centering
\caption{Evaluation results of the ablation study in the PT course. The results report Precision (PR), Recall (RE), and F1-measure (F1). Bold numbers indicate the highest value for each metric in each week.}
\label{table_expt2}
\resizebox{\textwidth}{!}{%
\begin{tabular}{|cc|ccc|ccc|ccc|ccc|ccc|ccc|ccc|}
\hline
\multirow{2}{*}{Dataset}    & \multirow{2}{*}{Week (s)} & \multicolumn{3}{c}{$\mathcal{L}_{CV}$+$\mathcal{L}_{HD}$+$\mathcal{L}_{KD}$ (Soft)} & \multicolumn{3}{c}{$\mathcal{L}_{CV}$+$\mathcal{L}_{KD}$ (Soft)} & \multicolumn{3}{c}{$\mathcal{L}_{HD}$+$\mathcal{L}_{KD}$ (Soft)} & \multicolumn{3}{c}{$\mathcal{L}_{CV}$+$\mathcal{L}_{HD}$} & \multicolumn{3}{c}{Only $\mathcal{L}_{CV}$} & \multicolumn{3}{c}{Only $\mathcal{L}_{HD}$} & \multicolumn{3}{c|}{Only $\mathcal{L}_{KD}$ (Soft)} \\ \cline{3-23}
&  & PR  & RE  & F1  & PR & RE & F1  & PR  & RE  & F1       & PR      & RE     & F1     & PR      & RE      & F1      & PR      & RE      & F1      & PR         & RE        & F1        \\ \hline
\multirow{4}{*}{T19P20}     & 1-3                       & 0.41          & \textbf{0.83} & 0.55          & 0.41          & 0.82          & 0.55          & 0.43          & \textbf{0.83} & 0.56          & 0.42          & \textbf{0.83} & 0.55          & 0.41          & 0.79          & 0.54          & 0.42          & 0.82          & 0.55          & \textbf{0.44} & 0.82 & \textbf{0.57} \\
                            & 1-4                       & 0.48          & \textbf{0.81} & 0.60          & 0.48          & 0.79          & 0.60          & \textbf{0.50} & 0.79          & \textbf{0.61} & \textbf{0.50} & 0.80          & \textbf{0.61} & 0.47          & 0.77          & 0.58          & 0.49          & 0.80          & \textbf{0.61} & 0.48          & 0.75 & 0.58          \\
                            & 1-5                       & 0.62          & 0.72          & 0.66          & 0.57          & \textbf{0.74} & 0.64          & 0.61          & 0.72          & 0.66          & \textbf{0.64} & 0.72          & \textbf{0.67} & 0.57          & \textbf{0.74} & 0.64          & 0.60          & \textbf{0.74} & 0.67          & 0.60          & 0.69 & 0.64          \\
                            & 1-6                       & 0.66          & 0.79          & 0.72          & 0.68          & 0.76          & 0.72          & \textbf{0.69} & 0.80          & \textbf{0.74} & 0.66          & 0.79          & 0.72          & 0.66          & 0.76          & 0.70          & \textbf{0.69} & \textbf{0.81} & \textbf{0.74} & \textbf{0.69} & 0.76 & 0.72          \\ \hline
\multirow{4}{*}{T20P21}     & 1-3                       & 0.56          & \textbf{0.47} & 0.50          & \textbf{0.60} & 0.43          & 0.50          & 0.55          & 0.42          & 0.47          & 0.56          & \textbf{0.47} & \textbf{0.51} & 0.58          & 0.44          & 0.49          & 0.53          & 0.44          & 0.47          & 0.58          & 0.37 & 0.44          \\
                            & 1-4                       & 0.59          & 0.44          & 0.50          & 0.59          & \textbf{0.45} & \textbf{0.51} & 0.56          & 0.40          & 0.46          & 0.59          & 0.43          & 0.49          & \textbf{0.60} & 0.44          & 0.50          & 0.56          & 0.41          & 0.47          & 0.58          & 0.41 & 0.47          \\
                            & 1-5                       & 0.71          & 0.47          & 0.55          & 0.71          & 0.47          & \textbf{0.56} & 0.71          & 0.41          & 0.52          & 0.69          & \textbf{0.48} & \textbf{0.56} & \textbf{0.73} & 0.44          & 0.55          & 0.70          & 0.46          & 0.55          & 0.72          & 0.43 & 0.53          \\
                            & 1-6                       & 0.83          & \textbf{0.52} & 0.64          & 0.78          & 0.47          & 0.58          & \textbf{0.86} & \textbf{0.52} & \textbf{0.65} & 0.81          & 0.51          & 0.62          & 0.78          & 0.49          & 0.60          & 0.85          & 0.51          & 0.63          & 0.78          & 0.46 & 0.57          \\ \hline
\multirow{4}{*}{T21P22}     & 1-3                       & 0.67          & 0.40          & 0.50          & 0.67          & 0.38          & 0.49          & 0.66          & \textbf{0.41} & \textbf{0.51} & \textbf{0.68} & 0.40          & 0.50          & 0.67          & 0.39          & 0.49          & 0.65          & 0.39          & 0.49          & \textbf{0.68} & 0.38 & 0.49          \\
                            & 1-4                       & 0.69          & 0.41          & 0.51          & 0.69          & 0.41          & 0.51          & 0.69          & 0.42          & 0.52          & 0.70          & \textbf{0.45} & \textbf{0.54} & 0.69          & 0.41          & 0.51          & 0.69          & 0.42          & 0.51          & \textbf{0.72} & 0.38 & 0.49          \\
                            & 1-5                       & \textbf{0.82} & 0.44          & \textbf{0.57} & 0.78          & 0.39          & 0.51          & 0.81          & 0.44          & 0.57          & 0.81          & 0.44          & \textbf{0.57} & 0.79          & 0.41          & 0.53          & 0.79          & \textbf{0.46} & \textbf{0.57} & 0.79          & 0.41 & 0.53          \\
                            & 1-6                       & 0.86          & 0.42          & 0.56          & \textbf{0.89} & 0.37          & 0.52          & 0.86          & 0.42          & 0.56          & 0.87          & 0.43          & \textbf{0.58} & 0.86          & 0.37          & 0.51          & 0.83          & \textbf{0.44} & 0.57          & 0.86          & 0.36 & 0.51          \\ \hline
\multirow{4}{*}{T1920P21}   & 1-3                       & \textbf{0.61} & 0.37          & 0.45          & 0.60          & \textbf{0.41} & \textbf{0.48} & 0.60          & 0.40          & 0.47          & 0.58          & 0.38          & 0.45          & 0.58          & \textbf{0.41} & 0.47          & 0.60          & 0.38          & 0.46          & 0.60          & 0.37 & 0.45          \\
                            & 1-4                       & \textbf{0.72} & \textbf{0.49} & \textbf{0.58} & 0.70          & 0.46          & 0.55          & 0.68          & 0.45          & 0.54          & 0.70          & 0.46          & 0.55          & 0.66          & 0.45          & 0.53          & 0.66          & 0.41          & 0.50          & 0.69          & 0.41 & 0.51          \\
                            & 1-5                       & 0.74          & \textbf{0.50} & \textbf{0.59} & \textbf{0.75} & 0.49          & \textbf{0.59} & 0.73          & 0.49          & 0.58          & 0.74          & 0.49          & 0.58          & 0.72          & 0.49          & 0.57          & 0.73          & 0.49          & 0.58          & \textbf{0.75} & 0.44 & 0.55          \\
                            & 1-6                       & 0.79          & 0.48          & 0.59          & 0.78          & \textbf{0.50} & 0.60          & 0.82          & 0.49          & 0.61          & 0.79          & 0.48          & 0.59          & 0.77          & 0.49          & 0.60          & \textbf{0.84} & 0.49          & \textbf{0.61} & 0.81          & 0.46 & 0.58          \\ \hline
\multirow{4}{*}{T2021P22}   & 1-3                       & \textbf{0.68} & 0.42          & 0.52          & 0.66          & 0.41          & 0.51          & 0.66          & 0.42          & 0.51          & 0.68          & \textbf{0.44} & \textbf{0.53} & 0.68          & 0.39          & 0.49          & 0.67          & 0.42          & 0.51          & 0.68          & 0.37 & 0.48          \\
                            & 1-4                       & 0.72          & 0.46          & 0.56          & \textbf{0.74} & 0.44          & 0.54          & 0.72          & 0.45          & 0.55          & 0.71          & 0.46          & 0.55          & \textbf{0.74} & \textbf{0.47} & \textbf{0.57} & 0.68          & 0.43          & 0.52          & 0.72          & 0.44 & 0.54          \\
                            & 1-5                       & 0.82          & 0.47          & 0.59          & 0.81          & 0.47          & 0.59          & \textbf{0.83} & 0.47          & 0.60          & 0.82          & 0.46          & 0.59          & 0.82          & 0.45          & 0.58          & \textbf{0.83} & \textbf{0.51} & \textbf{0.63} & 0.79          & 0.46 & 0.57          \\
                            & 1-6                       & 0.82          & 0.39          & 0.52          & 0.79          & 0.39          & 0.52          & \textbf{0.83} & 0.37          & 0.51          & 0.79          & \textbf{0.40} & \textbf{0.53} & 0.81          & 0.37          & 0.50          & 0.80          & 0.36          & 0.49          & 0.79          & 0.37 & 0.50          \\ \hline
\multirow{4}{*}{T192021P22} & 1-3                       & 0.74          & \textbf{0.43} & \textbf{0.54} & 0.73          & \textbf{0.43} & \textbf{0.54} & 0.71          & 0.40          & 0.50          & 0.75          & 0.42          & 0.54          & 0.74          & \textbf{0.43} & \textbf{0.54} & 0.70          & 0.41          & 0.51          & \textbf{0.77} & 0.36 & 0.49          \\
                            & 1-4                       & 0.76          & 0.43          & 0.55          & 0.77          & 0.42          & 0.54          & 0.74          & 0.43          & 0.54          & 0.76          & 0.43          & 0.55          & \textbf{0.80} & \textbf{0.44} & \textbf{0.56} & 0.75          & 0.43          & 0.54          & 0.75          & 0.42 & 0.53          \\
                            & 1-5                       & 0.83          & 0.54          & 0.65          & 0.81          & 0.50          & 0.61          & \textbf{0.84} & \textbf{0.58} & \textbf{0.68} & \textbf{0.84} & 0.53          & 0.65          & 0.83          & 0.49          & 0.60          & 0.80          & 0.57          & 0.66          & 0.81          & 0.46 & 0.59          \\
                            & 1-6                       & 0.81          & \textbf{0.48} & \textbf{0.60} & 0.80          & 0.44          & 0.56          & 0.80          & 0.46          & 0.58          & 0.79          & 0.47          & 0.59          & \textbf{0.82} & 0.45          & 0.58          & 0.78          & 0.46          & 0.58          & 0.80          & 0.43 & 0.56          \\ \hline
\end{tabular}%
}
\end{table*}

Table \ref{table_expt1} compares RNN-Attention-KD with several conventional algorithm baselines: MLP, RNN, GRU, LSTM, the bidirectional gated recurrent unit (Bi-GRU), and Bi-LSTM. RNN-Attention-KD is trained according to the KD strategy described in Section \ref{subsubsec322}, using the teacher model (RNN-Attention), with details in Table \ref{table_teachers}.

The baseline models are trained using SRP inputs directly from the first week to the predicted week. When students enroll in the PT course during Weeks 1--2, they are only exposed to a small portion of the course content, and the data collected during this period does not fully capture important behavioral patterns and learning progress. This limited scope of data makes it harder for prediction models to represent students' learning behavior and predict at-risk accurately. We thus begin identifying at-risk students starting from Weeks 3--6 since students engage with more content and provide more affluent and informative feature representations. It is important to note that students from the Faculty of Information Science and Electrical Engineering, Kyushu University can withdraw from a course only after five weeks of attending it. Therefore, instructors can use our framework to aid at-risk students before they withdraw.

Notably, the teacher model on the T192021P22 dataset yielded lower results than other datasets, even though the models were trained on the datasets from three PT courses. Three PT courses (PT2019, PT2020, and PT2021) introduce diverse students' learning patterns in both on-site and online classes. Students' learning patterns on PT2022 may have a contextual mismatch with previous data courses, diluting the relevance of older data to future courses and making it harder for the model on the T192021P22 dataset to learn stable patterns that generalize.

\begin{table}[t]
\caption{Performance metrics of the teacher model (RNN-Attention) trained on the entire PT course, used to train RNN-Attention-KD.}
\label{table_teachers}
\begin{tabular}{lccc}
\hline
Dataset    & Precision & Recall & F1-measure \\ \hline
T19P20     & 0.77      & 0.81   & 0.79       \\
T20P21     & 0.86      & 0.63   & 0.73       \\
T21P22     & 0.94      & 0.52   & 0.67       \\
T1920P21   & 0.92      & 0.63   & 0.75       \\
T2021P22   & 0.86      & 0.62   & 0.72       \\
T192021P22 & 0.77      & 0.69   & 0.73       \\ \hline
\end{tabular}
\end{table}

The results from Table \ref{table_expt1} show that RNN-Attention-KD outperforms conventional neural network models with high recall and F1-measure values in four out of six datasets: T20P21, T21P22, T2021P22, and T192021P22. Notably, the high recall values mean our model could identify most at-risk students from the total population of at-risk students in early states better than the conventional algorithm model. Although RNN-Attention-KD performed better, it fell short of GRU, Bi-GRU, and Bi-LSTM in the T19P20 and T1920P21 datasets. The reason may be attributed to the fact that the PT2019 course was conducted as an on-site class, and PT2020, PT2021, and PT2022 were online classes due to the COVID-19 pandemic. The difference in learning environments likely affected students' learning patterns, which could explain the decrease in performance when training with on-site data and predicting with online course data. RNN-Attention-KD achieved the highest average recall and F1-measure across all datasets, outperforming other models. For example, it obtained recall and F1-measure of 0.49 and 0.51 for Weeks 1--3 and 0.51 and 0.61 for Weeks 1--6, respectively.

In addition, RNN models are designed to leverage the temporal context and patterns within time-sequence data. Since removing the later week of data may eliminate an important portion of sequential information that might have been crucial for making accurate predictions, predictive performance was decreased on student models on all datasets. In other words, the additional week of students' learning data enabled the RNN to identify patterns better and ultimately make more accurate classifications, which explains why the F1-score dropped when those data were removed. 

Notably, the time-dependent algorithms, such as RNN, GRU, and LSTM, provided better performance than MLP, which is a time-independent algorithm, on all metrics. This demonstrates an important property of time-dependent algorithms -- the ability to capture the relationships between sequential data points and consider how data evolves at each time step \cite{lopez2022}. 

However, the grade distribution in PT courses is somewhat skewed, especially in PT2020 and PT2021, as shown in Table \ref{table_grade}. This imbalance influences the prediction model's learned decision boundaries. In addition, this may make the prediction models struggle to recognize students as non-at-risk more than at-risk students, which affects the recall and F1-score due to insufficient examples. To mitigate these biases, we will adjust the model, apply oversampling techniques, and collect more features to obtain more representative data, which may be necessary to mitigate possible biases and achieve more consistently reliable predictions.

\subsection{Ablation Study}
\label{subsec52}

Next, we conduct ablation studies on RNN-Attention-KD to examine the contributions of different distillation objectives:  hint loss $ \mathcal{L}_{HD} $ (Equation \ref{loss_hidden}),  context vector loss $ \mathcal{L}_{CV} $ (Equation \ref{loss_context_vector}), and distillation loss on the soft cross-entropy loss $ \mathcal{L}_{KD} $ (Equation \ref{loss_distillation}).

The ablation study results are listed in Table \ref{table_expt2}. RNN-Attention-KD utilizing the knowledge that distills from all objectives and using only the hint and context loss did not show significant performance differences. Therefore, we suggest that RNN-Attention-KD, which used only two components—hint loss and context vector loss—is sufficient to transfer knowledge from the teacher model to the student model to improve the prediction performance. This can simplify the distillation process and reduce the computational complexity, which makes the training process more efficient.

On the contrary, the model that relied solely on the distillation loss on the soft cross-entropy loss from the teacher model had the worst recall and F1-score. This means the soft cross-entropy loss on the distillation loss was an obstacle to the performance on RNN-Attention-KD. Soft cross-entropy loss transfers a finer knowledge called ``dark knowledge'' from predicted logits of at-risk and no-risk classes on the output units on the teacher to the student model. Soft cross-entropy loss might force the student model to learn unnecessarily complex patterns, leading to overfitting that makes a student model perform poorly. Even though we set a lower weight of hyperparameter $ \lambda $ (= 0.1) that uses soft and hard cross-entropy losses for balances, the ability to generalize on RNN-Attention-KD was degraded by focusing too much on approximating the predicted logits on the teacher model rather than learning from the ground truth labels. Therefore, the hint loss and the context vector loss can enhance model effectiveness for identifying at-risk individuals.


\section{Conclusions and Future Work}
\label{sec6}

This study introduced RNN-Attention-KD, a novel framework for identifying at-risk students early. This is a crucial task in EDM research to prevent students from dropping out, enabling instructors to provide timely support. RNN-Attention-KD employed RNNs that can handle time-series information in educational data. It also leveraged the benefits of attention mechanisms to solve the shortcomings of RNNs that had a vanishing gradient when capturing long-term time series, retaining vital information by focusing on critical parts of the input sequence. KD is used to compress time series to identify at-risk students by distilling the knowledge from the teacher model, which was trained on the entire course data to increase model prediction performance.

The empirical results showed that RNN-Attention-KD outperforms traditional neural network models with high recall and F1-measure values. Nevertheless, the case that utilized on-site classes as a training dataset and online classes as a test set decreased the RNN-Attention-KD performance since students’ learning patterns differed in these different learning environments and modalities. 

In addition, an ablation study was conducted to investigate the contributions of different distillation objectives. We found that hint loss from the hidden layer and context vector loss from the attention module on RNN can enhance prediction performance more than the distillation loss on the soft cross-entropy loss from the teacher model. The last case led to overfitting and thus making a student model perform poorly.

This study has some limitations that remain as open challenges to be addressed in future research.
\begin{enumerate}
    \item Due to the data being collected from only one subject at one university, the model's adaptability across different learning environments still needs to be investigated. Future research should examine the model's performance and tuning strategies in various instructional contexts, such as different subject areas, educational levels, and teaching methods. Additionally, we will explore how the model can adapt through machine learning techniques, such as domain adaptation, to enhance its generalizability.
    \item The present study was not conducted on deploying the RNN-Attention-KD for early prediction within actual classes. The model's usability in real-world applications still remains to be evaluated. Future research should conduct studies in actual classes to investigate model adaptation and refinement strategies. This can improve the model's generalizability based on instructor feedback and provide richer insights into how to support at-risk students' learning early.
\end{enumerate}

In addition to the above, future research may improve the generalizability and performance of the RNN-Attention-KD model using the multi-teacher models. Each teacher model has specific knowledge of each student's learning log data, such as reading behaviors, course assessment, and assignment data. This will make the prediction performance of the prediction model more robust.

\begin{acks}
This work was supported by JST CREST Grant Number JPMJCR22D1 and JSPS KAKENHI Grant Number JP22H00551, Japan.
\end{acks}

\appendix

\section{Supplementary Materials}
The code written to produce the results reported in this study is publicly available at: \url{https://github.com/limu-research/2025-SAC-RNN-Attention-KD}. This enables other researchers to more easily adopt or adapt our methods.